\documentclass[lettersize,journal]{IEEEtran}
\usepackage{xcolor}
\usepackage{graphicx}
\usepackage{hyperref}
\usepackage{pgfplots}
\usepackage{pgfplotstable}
\pgfplotsset{compat=1.17}
\usepackage{array}
\usepackage{booktabs}
\usepackage{amssymb}
\usepackage{comment}

\ifCLASSINFOpdf
\else
\fi
\begin{document}
%
\title{ AIR-VIEW: The Aviation Image Repository for Visibility Estimation of Weather, A Dataset and Benchmark}

%
%
%

\author{Chad Mourning,~\IEEEmembership{Member,~IEEE,}
        Zhewei Wang,~\IEEEmembership{Member,~IEEE,}
        Justin Murray,~\IEEEmembership{Student Member,~IEEE}
\thanks{Chad Mourning, Zhewei Wang, and Justin Murray are with the School
of Electrical Engineering and Computer Science, Ohio University, Athens, OH 45701}}

%
%

\markboth{Journal of \LaTeX\ Class Files,~Vol.~14, No.~8, August~2015}%
{Shell \MakeLowercase{\textit{et al.}}: Bare Demo of IEEEtran.cls for IEEE Journals}
%



\maketitle

\begin{abstract}
Machine Learning for aviation weather is a growing area of research for providing low-cost alternatives for traditional, expensive weather sensors; however, in the area of atmospheric visibility estimation, publicly available datasets, tagged with visibility estimates, of distances relevant for aviation, of diverse locations, of sufficient size for use in supervised learning, are absent.  This paper introduces a new dataset which represents the culmination of a year-long data collection campaign of images from the FAA weather camera network suitable for this purpose.  We also present a benchmark when applying three commonly used approaches and a general-purpose baseline when trained and tested on three publicly available datasets, in addition to our own, when compared against a recently ratified ASTM standard.
\end{abstract}


%
\IEEEpeerreviewmaketitle

\section{Introduction}

\IEEEPARstart{P}{oor} visibility is a leading cause of aircraft accidents, particularly so called ``VFR-into-IMC" where pilots go from conditions meeting the criteria for \emph{Visual Flight Rules} to \emph{Instrument Meteorological Conditions} \cite{faa-cfr}.  The leading atmospheric properties that drive this determination are \emph{atmospheric visibility}, a horizontal measurement, and its vertical dual, \emph{ceiling height}.

There are a handful of relevant definitions of atmospheric visibility including the World Meteorological Organization (WMO), Federal Aviation Administration (FAA), and International Civil Aviation Organization (ICAO).  The WMO definition is physically focused calling it ``the length of path in the atmosphere required to reduce the luminosity of a collimated beam to 5\% of its original value" \cite{wmo_no_8}, whereas the FAA and ICAO definitions are, predictably, aviation focused.  The FAA defines visibility as the ``measure of the horizontal opacity of the atmosphere at the point of observation and is expressed in terms of the horizontal distance at which a person should be able to see and identify a specific object" \cite{FAA7900.5} and ICAO is more precise defining visibility as ``the greatest distance at which a black object of suitable dimensions, situated near the ground, can be seen and recognized when observed against a bright background", or, alternatively, ``the greatest distance at which lights in the vicinity of 1,000 candelas can be seen and identified against an unlit background" \cite{icao_annex3}.

Existing methods for estimating these visibility values require expensive hardware with a small geographic footprint, and research into creating low-cost Machine Learning based estimators has started to gain momentum.  However, a true generalized solution for aviation has yet to be created, in part due to the lack of a diverse, high-quality, truth tagged imagery.

The main contributions of this paper are:

\begin{enumerate}
    \item Introduction of a new dataset to assist in training Machine Learning algorithms attempting to estimate atmospheric visibility.
    \item A benchmark representing the effectiveness of current approaches on estimating visibility when trained on this and other publicly available datasets.
\end{enumerate}

Section \ref{sec:problem} will elaborate on the visibility estimation problem and present some related work that has been done trying to solve it.  Section \ref{sec:datasets} covers related publicly available datasets researchers may use to train their models and how they differ from the one presented in this paper.  Section \ref{sec:dataset} presents some descriptive statistics about the dataset and Section \ref{sec:benchmark} presents the results of our benchmark using various methods on our dataset.  We then finish with some discussion on limitations of these approaches and possible future research directions.

\section{The Visibility Problem}
\label{sec:problem}

Estimating atmospheric visibility is not only of interest in aviation.  Some of the earliest attempts at estimating atmospheric visibility were among 18\textsuperscript{th} century astronomers attempting to correct for attenuated starlight, leading to mathematical models like Beer's Law and the Beer-Lambert Law \cite{beer}.  More recently, it has also been of substantial interest to developers of self-driving cars; however, the distances required for pilots, frequently 10 miles or more, require much more robust approaches than for ground vehicles.

Additionally, much of the focus in aviation on visibility estimation is centered around airports.  The current ``gold standard" for visibility estimates is the Automated Weather Observation System (AWOS) \cite{awos}, which uses transmissometer or scatterometry to infer an extinction coefficient of light and therefore a visibility estimate.  This can be seen by the FAA's large investment in weather cameras near airports and their partnership with MIT Lincoln Lab to develop the VEIA \cite{matthews2023veia} algorithm.  This highlights a dichotomy in visibility estimation research: full-reference vs. no-reference estimations.

VEIA requires a clear day reference image, which it infers as the clearest image over a few days from deployment, where it finds the mean edge weight using classical computer vision techniques; it then finds the mean edge weight in a current image and derives visibility.  Our work aims to support a no-reference model where an image taken from a cell-phone has just as much likelihood to get a good visibility estimate as a one million dollar FAA weather camera installation.

Li et al.'s Integrated Model \cite{li} were one of the first to attack this problem; however, their accuracies only reached 61\%. RMEP \cite{RMEP} is a regression based approach which achieved an RSME of around 1 kilometer, which is within FAA guidelines; however, they trained and tested on images near 11 AM.  While this partially controls irradiance in images, it raises questions about the generalness of their approach.

VisNet \cite{visnet} is a leading contender for visibility estimation, with the original paper classifying distances up to 20km (over 12 miles).  VisNet's key insight was the use of an Fourier Transform pre-processing step to isolate high-frequency noise as a proxy for loss of visibility in images.  Unfortunately, the VisNet authors did not publicly release their dataset, so we have been unable to verify their results. We also have concerns about the lack of adaptability in the FFT spectra, which may not generalize to varying resolution cameras.

It is worth considering if popular, off-the-shelf models are effective at the visibility estimation task; Kevin Kipfer \cite{kipfer2017fog} compared various well known architectures where VGCC-16 \cite{simonyan2014very}  achieved the highest accuracy at 52\%.  Similarly we test a ResNet50 \cite{koonce2021resnet} model against the new dataset.


\section{Related Datasets}
\label{sec:datasets}

This section describes some of the popular datasets currently used by researchers and how our new dataset compares. 
 There are a handful of issues which make various datasets less than ideal for aviation-relevant visibility estimation, in a rough order of importance:

 \begin{enumerate}
    \item Visibility values - if the dataset does not have tagged visibility estimates, value in supervised learning models is greatly reduced.
     \item Number of images - without sufficient data Machine Learning models will fail to generalize.
     \item Open Access - a control implementation is crucial in science, and without access to previous papers' datasets it is difficult to quantify advancement.
     \item Variety in scenes - for robust generalization diverse locations are preferred, otherwise overfitting to the terrain or other features in locations may occur.
     \item Real images - while sufficiently realistic synthetic images may train models that generalize to reality, sim-to-real transfer in this problem has little validation as is an ongoing area of study.
 \end{enumerate}

The FRIDA\footnote{\url{http://perso.lcpc.fr/tarel.jean-philippe/bdd/frida.zip}} (Foggy Road Image DAtabase) and FRIDA2\footnote{\url{http://perso.lcpc.fr/tarel.jean-philippe/bdd/frida2.zip}} datasets are synthetic datasets with 90 and 330 images, respectively, of 18 scenes from 2010, but no tagged visibility distances.  The Foggy Cityscapes and Foggy Driving datasets\footnote{\url{https://people.ee.ethz.ch/~csakarid/SFSU_synthetic/}} from \cite{SDV18} contain nearly 25,000 synthetic images with tagged visibilities up to 600m and 500 real-images with no visibility estimates.  The FRIDA and Foggy Cityscapes datasets provide per-pixel depth map information as well, which may be useful for training self-driving cars, but is not well suited for atmospheric visibility estimation.  SynFog \cite{xie2024synfog} is a recent advance in synthetic datasets, but, again, focuses on self-driving vehicle distances and have yet to release their dataset (stated as ``coming soon").

The ``Foggy Road Sign" dataset (FORSI) \cite{belaroussi2014impact} contains around 3500 synthetic images, is publicly available, but only has visibilities up to 250 meters.  Schaupp et. al \cite{schaupp2025synthetic} have ongoing research on high-fidelity synthetic imagery for visibility estimation using various fog models, at avaiation relevant distances, but no dataset has been released.

The FVEI (Fog Visibility Estimation Image) dataset introduced in \cite{yang2023multi} is slightly better containing 15,000 images visibility tagged images, but distances are all 500m or less and appears to be gated behind registration on Baidu Pan a Chinese cloud storage service.  The DENSE dataset\footnote{\url{https://www.uni-ulm.de/en/in/institute-of-measurement-control-and-microtechnology/research/data-sets/dense-datasets/}} \cite{gated2depth2019} is particularly interesting because it is based on LIDAR through fog, another automotive use case, is publicly available, after registration, but only provided automotive relevant distances.

FOVI (Foggy Outdoor Visibility Images), the dataset of used by the authors of VisNet \cite{visnet}, which the authors state contained 3,000,000 images from 26 locations in South Korea.  This is a remarkable number of images, and a reasonably diverse set of locations; it also contains distances relevant to aviation applications, unfortunately, this dataset is not publicly available. The massive Archive of Many Outdoor Scene (AMOS)\footnote{\url{https://mvrl.cse.wustl.edu/datasets/amos/}} \cite{jacobs07amos} datasets contain an abundance of images and a diversity of locations, 17 million from 538 sites, but they lack visibility truth values.

RMEP uses the SSF (Sub-SkyFinder) \cite{RMEP} dataset, downselected to 24,000 images from 30 sites, retroactively tagging them with records of visibility at their location and time is a publicly available dataset we evaluated in this paper.  This is probably the most well-suited dataset to the atmospheric visibility estimation task before the release of our dataset.

\section{The Dataset}
\label{sec:dataset}

In January of 2024, while presenting our preliminary work on the visibility estimation problem \cite{mourning2024towards}, it was recommended that we consider leveraging the FAA Weather Camera network as a source of training data for their model.  After contacting the administrator to solicit bulk data, we were informed that the FAA does not keep images from the cameras after they are provided to their users, and that they do not provide an API to collect these images programmatically.

After this discussion, we embarked on a data collection campaign to scrape images from the various FAA weather camera sites that are co-located with visibility sensors to create a truth dataset for the visibility estimation problem.  The following section describes our dataset.

\subsection{Descriptive Statistics}

\begin{figure}

\begin{tikzpicture}
\begin{axis}[
ybar,
    bar width=5pt,
    xlabel=Visibility,
    ylabel=Count,
    xtick={1,2,3,4,5,6,7,8,9,10,11},
    xticklabels={$\!<$1,1,2,3,4,5,6,7,8,9,10+},
    width=\columnwidth,
    height=8cm,
    enlargelimits=0.05,
    grid=major,
    ymode=log,
]
\addplot table[x index=0, y index=2, col sep=comma] {visibility_counts.csv};
\end{axis}
\end{tikzpicture}

\caption{A histogram of the visibilities found in the dataset.  Values less than one are combined into one bin, but still only total 28 instances for the entire dataset.  This is a logarithmic plot, so instances of 9 and 10+ miles of visibility dominate the data.}
\label{fig:visibilities}
\end{figure}
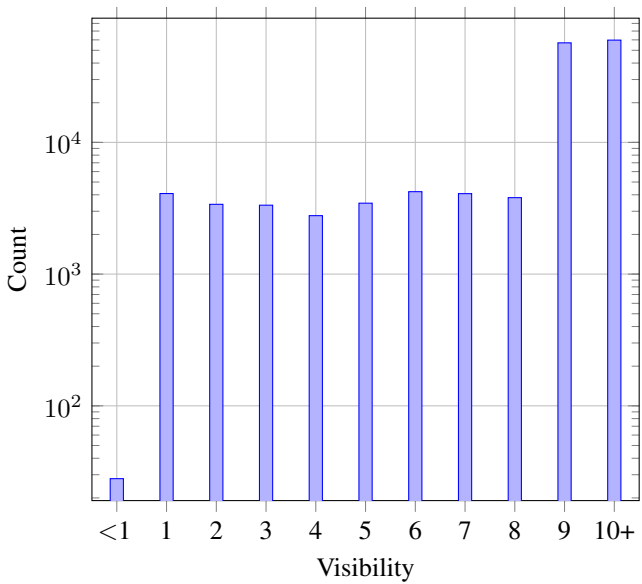

Version 1 of AIR-VIEW contains 147,182 images, collected between April 9\textsuperscript{th}, 2024 and May 17\textsuperscript{th}, 2025 from 444 different FAA weather camera sites; only sites with visibility estimates are included.  First, we present in Figure \ref{fig:visibilities}, the key feature of the dataset, the visibility truth values.  Valid visibilities for automated reports can be found in \cite{awos}; however, for clarity, partial mile visibilities are rolled together in the histogram.  Most days are clear, 9 and 10+ mile samples make up 80\% of the dataset, so users should take care to re-balance the data.

\begin{figure}

\begin{tikzpicture}
\begin{axis}[
ybar,
    bar width=5pt,
    xlabel=Harvests Per Month,
    ylabel=Count,
     ylabel near ticks,
    xtick={1,2,3,4,5,6,7,8,9,10,11,12},
    xticklabels={Jan, Feb, Mar, Apr, May, Jun, Jul, Aug, Sep, Oct, Nov, Dec},
    xticklabel style={rotate=0},
    width=\columnwidth,
    height=8cm,
    enlargelimits=0.05,
    grid=both
]
\addplot table[x index=0, y index=2, col sep=comma] {day_counts.csv};
\end{axis}
\end{tikzpicture}

\caption{A histogram of the number of harvests per month.  In general, there were more harvest during months with more daylight.  An unexpected system outage led to only one harvest during November 2024. }
\label{fig:months}
\end{figure}
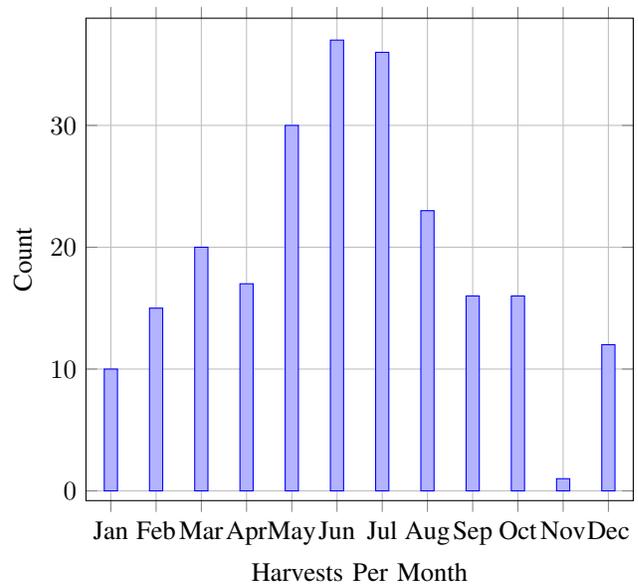

A scheme was devised to harvest FAA weather camera images at random intervals to provide meaningful variety in time-of-day in the dataset when taken in conjunction to the longitudinal span of the camera sites; however, early harvests on the east coast of the United States may still have been before sunrise in Alaska, especially during winter months.  Figure \ref{fig:months} shows the number of harvests per month; one will notices an increase in valid harvests during summer months with additional sunlight.  An error in the harvesting scheme led to only a single pull in November, but we believe the dataset contains enough meaningful winter images to be useful.

\begin{figure}
    \centering
    \includegraphics[width=0.99\linewidth]{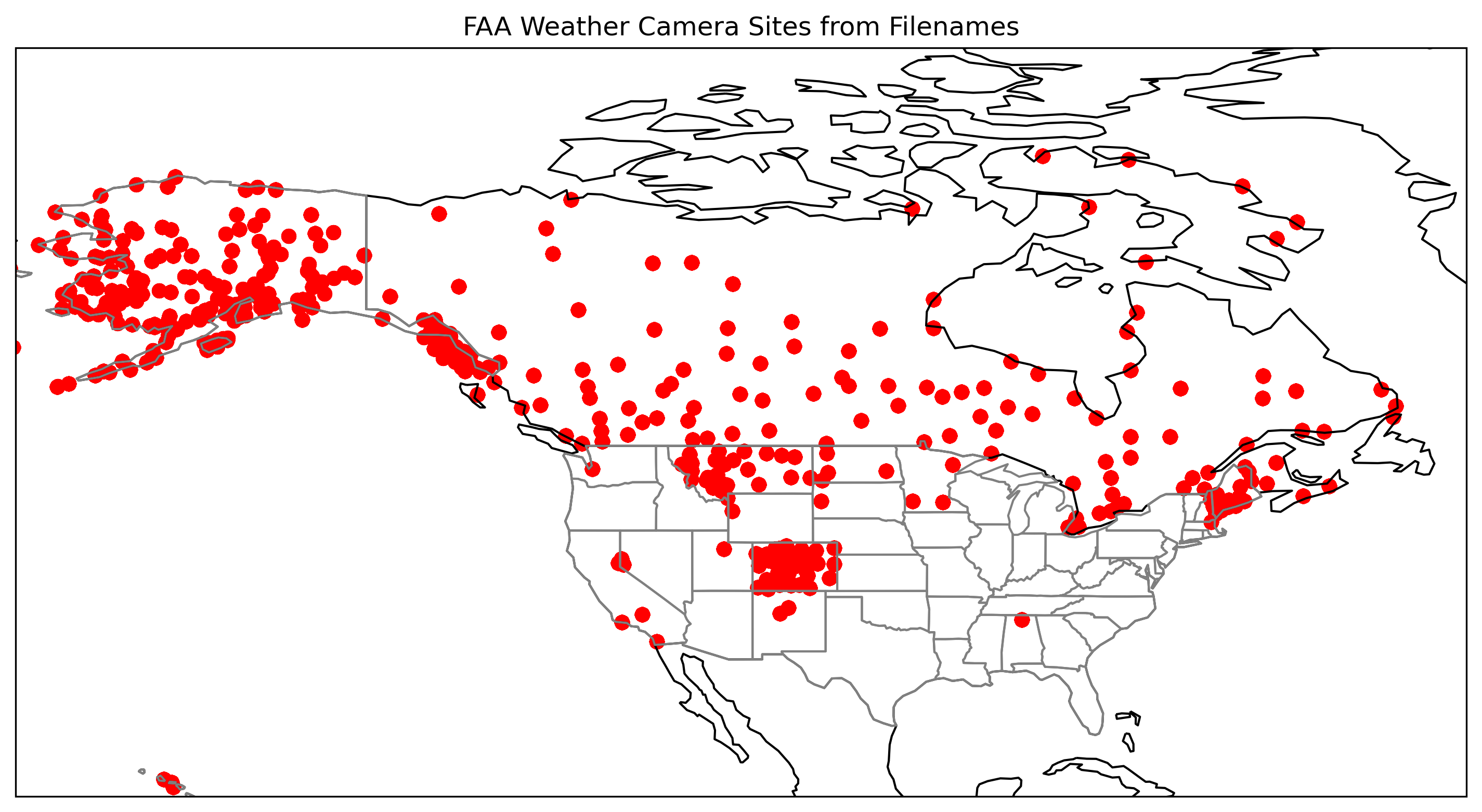}
    \caption{Locations of the subset of the 444 FAA Weather Cameras that included co-located visibility estimates harvested for our dataset.}
    \label{fig:locations}
\end{figure}

Originally, images were pulled from all of the FAA Weather Camera sites, but only a subset of the FAA Weather Cameras give a visibility estimate on the website, so only those images were collected in the later stages of the campaign, and only those images are included in AIR-VIEWv1. Figure \ref{fig:locations} shows the locations of camera sites included in the dataset. Even though the weather cameras are accessed through an FAA website, there are many partnering cameras providing data in Canada.

\begin{figure}

\begin{tikzpicture}
\begin{axis}[
ybar,
    bar width=5pt,
    xlabel=Harvests Per Hour,
    ylabel=Count,
     ylabel near ticks,
    xtick={0,2,4,6,8,10,12,14,16,18,20,22},
    xticklabel style={rotate=0},
    width=\columnwidth,
    height=8cm,
    enlargelimits=0.05,
    grid=both,
]
\addplot table[x index=0, y index=1, col sep=comma] {local_hour_counts.csv};
\end{axis}
\end{tikzpicture}

\caption{A histogram of the number of harvests per hour-of-day.  Bins contain the number of harvests that contain a time within a given hour after adjusting for timezone.  For example, bin 8 contains the number of instances containing images collected between 8:00 AM and 8:59 AM in the local time zone of the image. Every even hour is labeled for clarity.}
\label{fig:hours}
\end{figure}
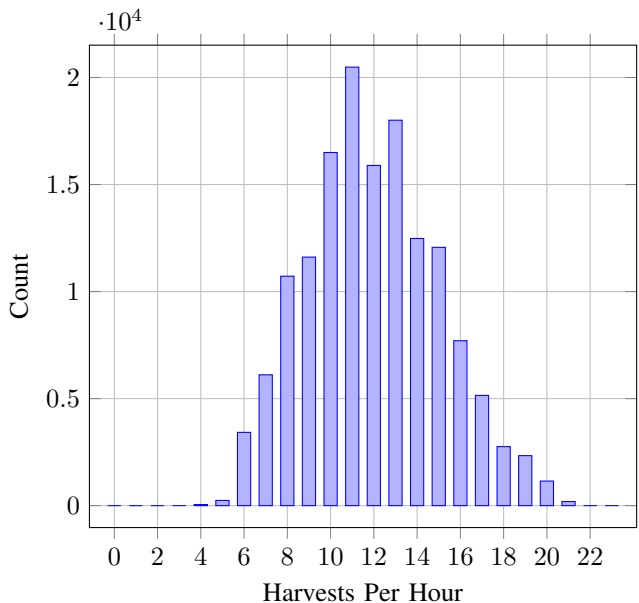

Another consideration is time of day.  The RMEP \cite{RMEP} model intentionally downselected to images around 11AM to attempt to control for irradiance in images; however, this raises concerns about generality.  Figure \ref{fig:hours} shows a histogram of hours of image captures.  The dataset can be found here\footnote{10.5281/zenodo.15742555}.

\section{The Benchmark}
\label{sec:benchmark}

\paragraph{Methodology}

To evaluate the effectiveness of our dataset against the other publicly available models, we will evaluate three popular visibility estimation architectures, VisNet \cite{visnet}, RMEP \cite{RMEP}, and Li's Integrated Model \cite{li}.  Additionally, as a baseline, we evaluate a popular general-purpose architecture ResNet50 \cite{koonce2021resnet}.

For the AIR-VIEW tests, we curate the dataset to balance for class representation, and train regression-style networks for better comparison against the ASTM standard for weather information providers \cite{ASTM_F3673-23}.  These results should serve as a starting point for any reseaerchers attempting to improve upon the state-of-the-art in atmospheric visibility estimation.



\paragraph{Results}

Table~\ref{tab:result} shows the results for each of the four models trained and tested on each of the three publicly available datasets, as well as trained on their own as tested against AIR-VIEW.  Each cell for the existing datasets contains typical regression metrics including: MAE, MAPE, MSE, as well as MRSE, when trained on the dataset from that column and tested against that dataset (left) and when tested against AIR-VIEWv1 (right).  The AIR-VIEW column only shows the value when trained and tests on AIR-VIEW, and includes an additional metric derived from the ASTM WIP standard \cite{ASTM_F3673-23} for Tier I data, translating to percentage of images within 1.5 miles of their correct value.

\begin{table*}[ht]

\centering
\renewcommand{\arraystretch}{1.5}
\setlength{\tabcolsep}{8pt}
\begin{tabular}{l|>{\raggedright\arraybackslash}p{3.2cm}
                >{\raggedright\arraybackslash}p{3.4cm}
                >{\raggedright\arraybackslash}p{3.2cm}
                >{\raggedright\arraybackslash}p{3cm}}
\toprule
\textbf{Method \textbackslash~Dataset} & \textbf{FROSI} & \textbf{FCS} &   \textbf{SSF} &\textbf{AIR-VIEWv1} \\
\midrule
\textbf{VisNet} \cite{visnet} &
MAE: 4.20 m / 4.34 mi\newline
MAPE: 2.06\% / 158\%\newline
MSE: 347 m$^{2}$ / 27.2 mi$^{2}$\newline
RMSE: 18.6 m / 5.22 mi\newline
&
MAE: 71.8 m / 5.02 mi\newline
MAPE: 24.5\% / 98.8\%\newline
MSE: 9,950 m$^{2}$ / 32.6 mi$^{2}$\newline
RMSE: 99.8 m / 5.71 mi\newline
&
MAE: 1.42 mi / 4.54 mi\newline
MAPE: 87.1\% / 196\%\newline
MSE: 7.02 mi$^{2}$ / 28.2 mi$^{2}$\newline
RMSE: 2.65 mi / 5.31 mi\newline
&
MAE: 1.83 mi\newline
MAPE: 56.5\%\newline
MSE: 5.56 mi$^{2}$\newline
RMSE: 2.38 mi\newline
ASTM: 75\%
\\
\midrule
\textbf{RMEP \cite{RMEP}} &
MAE: 5.12 m / 11.4 mi\newline
MAPE: 0.0331\% / 95.4\%\newline
MSE: 0.004 m$^{2}$ / 31.6 mi$^{2}$\newline
RMSE: 0.066 m / 5.62 mi\newline
&
MAE: 38.3 m / 4.86 mi\newline
MAPE: 13.4\% / 97\%\newline
MSE: 1 $\times 10^4$ m$^{2}$ / 32.3 mi$^{2}$\newline
RMSE: 102 m / 5.68 mi\newline
&
MAE: 1.35 mi / 4.66 mi\newline
MAPE: 60.1\% / 140\%\newline
MSE: 7.76 mi$^{2}$ / 19.0 mi$^{2}$\newline
RMSE: 2.78 mi / 4.36 mi\newline
&
MAE: 1.79 mi\newline
MAPE: 54.3\%\newline
MSE: 5.48 mi$^{2}$\newline
RMSE: 2.34 mi\newline
ASTM: 75\%\newline
\\
\midrule
\textbf{Li Integrated \cite{li}} &
MAE: 3.31 m / 3.62 mi\newline
MAPE: 2.22\% / 60.12\%\newline
MSE: 26.8 m$^{2}$ / 21.3 mi$^{2}$ \newline
RMSE: 5.18 m / 4.62 mi \newline
&
MAE: 50.8 m / 4.86 mi\newline
MAPE: 16.1\% / 97.2\%\newline
MSE: 5,930 m$^{2}$ / 32.2 mi$^{2}$ \newline
RMSE: 77.0 m / 5.67 mi \newline
&
MAE: 1.25 mi / 4.15 mi\newline
MAPE: 72.4\% / 180\%\newline
MSE: 6.37 mi$^{2}$ / 25.0 mi$^{2}$ \newline
RMSE: 2.52 mi / 5.0 mi \newline
&
MAE: \textbf{1.77 mi}\newline
MAPE: 56.2\%\newline
MSE: \textbf{5.25 mi$^{2}$}\newline
RMSE: \textbf{2.29 mi}\newline
ASTM: \textbf{77\%}
\\
\midrule
\textbf{ResNet50 \cite{koonce2021resnet}} &
MAE: 1.58 m / 4.75 mi\newline
MAPE: 0.789\% / 93.9\%\newline
MSE: 4.63 m$^{2}$ / 31.2 mi$^{2}$\newline
RMSE: 2.15 m / 5.58 mi\newline
&
MAE: 36.2 m / 4.81 mi\newline
MAPE: 12.6\% / 95.8\%\newline
MSE: 5,960 m$^{2}$ / 31.8 mi$^{2}$\newline
RMSE: 77.2 m / 5.64 mi\newline
&
MAE: 1.22 mi / 3.72 mi\newline
MAPE: 66.1\% / 158\%\newline
MSE: 6.68 mi$^{2}$ / 20.4 mi$^{2}$\newline
RMSE: 2.59 mi / 5.52 mi\newline
&
MAE: \textbf{1.77 mi}\newline
MAPE: \textbf{52.1\%}\newline
MSE: 5.46 mi$^{2}$\newline
RMSE: 2.34 mi\newline
ASTM: 73\%
\\
\bottomrule
\end{tabular}
\caption{Evaluation metrics for each method on each dataset.  The four datasets being evaluated represent the columns, and the four techniques being evaluated represent the rows.  Each cell has 5 metrics, and each metric in the first three columns has two values: the results when trained and tested on the dataset from that column, as well as trained on their dataset and tested on AIR-VIEW. The final column presents the result when the technique is trained and tested on AIR-VIEW.  The best values in the final column are bolded.  The final column has an additional metric ``ASTM" which represents the ability to hit Tier I data in the ASTM WIP standards \cite{ASTM_F3673-23}}
\label{tab:result}
\end{table*}

\paragraph{Discussion}

Given the large number of images compared to the other publicly available datasets and small number of distances, it is unsurprising that models trained on FROSI, FCS, and SSF do not generalize.  Surprisingly, ResNet does as well compared to task-tailored models.

While the oldest, the Li Integrated Model \cite{li} seems to be the best performer overall on this new dataset.  ResNet also performed well, outperforming two of the task-tailored model on this dataset.  In an unreleased study, the FAA Weather Technology in the Cockpit program used a 20\% MAPE as a benchmark for AWOS agreement with aggregate expert estimate of visibility images; none of these example reach that threshold, -- more research is required.

\paragraph{Limitations}

The main drawback of optical visibility estimation is that it struggles without a light source, such as during the night.   Due to the nature of the image harvesting process, some lowlight images exist in the dataset; early morning harvests on the East Coast may be before sunrise in Alaska, and evening pulls in Alaska may be after sunset on the East Coast.  These should be discluded from training.

\paragraph{Conclusions}

Each of the models generally perform well when trained and tested on the same data, but their tests on real data for models trained on FROSI and FCS are generally abysmal, with RSMEs around 5 miles indicating roughly random chance behaviors.  Models trained on the SSF dataset start to approach reality, but still struggle with the webcam dataset, this could also be attributed to camera parameter differences such as field-of-view, focal point, etc.

The authors would implore Palvanov et al. to release their dataset for testing new approaches to this problem, as we have not been able to recreate their impressive performance on our dataset.  Until that time, we believe our dataset is the best publicly available dataset truth tagged with co-located certified visibility measurement equipment.

\section{Future Work}

The collection campaign will continue.  Very low visibility images are still underrepresented in the dataset and we do not believe most models have reached saturation.  The FAA continues to deploy new cameras, which will be routinely added to the dataset.  We intend to release updated AIR-VIEW datasets on an annual basis.

The authors' next area of interest in visibility estimation is Distribution Shifts, which can also likely be studied from this dataset.  Seasonal models, monthly models, or models with a moving window may improve performance may be relevant.  The other research direction is optical estimation of ceiling height.  The authors expect this may not be solvable with a single perspective. Full sky imagers exist and may be able to be deployed to collect a useful dataset for training.

\section*{Acknowledgment}

The work was inspired, in part, by the Ohio Federal Research Network Project 502: The Low Altitude Weather Network.  Additionally, this work would not have been possible with the FAA Weather Camera Network.

\ifCLASSOPTIONcaptionsoff
  \newpage
\fi



%
\bibliographystyle{IEEEtran}

\begin{thebibliography}{10}
\providecommand{\url}[1]{#1}
\csname url@samestyle\endcsname
\providecommand{\newblock}{\relax}
\providecommand{\bibinfo}[2]{#2}
\providecommand{\BIBentrySTDinterwordspacing}{\spaceskip=0pt\relax}
\providecommand{\BIBentryALTinterwordstretchfactor}{4}
\providecommand{\BIBentryALTinterwordspacing}{\spaceskip=\fontdimen2\font plus
\BIBentryALTinterwordstretchfactor\fontdimen3\font minus
  \fontdimen4\font\relax}
\providecommand{\BIBforeignlanguage}[2]{{%
\expandafter\ifx\csname l@#1\endcsname\relax
\typeout{** WARNING: IEEEtran.bst: No hyphenation pattern has been}%
\typeout{** loaded for the language `#1'. Using the pattern for}%
\typeout{** the default language instead.}%
\else
\language=\csname l@#1\endcsname
\fi
#2}}
\providecommand{\BIBdecl}{\relax}
\BIBdecl

\bibitem{faa-cfr}
{Federal Aviation Administration}, ``{14 CFR § 91.155 - Basic VFR Weather
  Minimums},'' Code of Federal Regulations, Title 14, Part 91, Section 155,
  2021, {Accessed on April 29, 2023}.

\bibitem{wmo_no_8}
\BIBentryALTinterwordspacing
{World Meteorological Organization}, \emph{Guide to Instruments and Methods of
  Observation}, 8th~ed., ser. WMO-No. 8.\hskip 1em plus 0.5em minus 0.4em\relax
  Geneva, Switzerland: World Meteorological Organization, 2018, accessed:
  2024-11-24. [Online]. Available:
  \url{https://library.wmo.int/doc_num.php?explnum_id=4147}
\BIBentrySTDinterwordspacing

\bibitem{FAA7900.5}
{Federal Aviation Administration}, \emph{FAA Order JO 7900.5E: Surface Weather
  Observing—METAR and SPECI}, U.S. Department of Transportation, Washington,
  DC, 2023, available at: \url{https://www.faa.gov}.

\bibitem{icao_annex3}
{International Civil Aviation Organization}, \emph{Annex 3 to the Convention on
  International Civil Aviation: Meteorological Service for International Air
  Navigation}, 20th~ed., International Civil Aviation Organization, Montreal,
  Canada, 2018, iCAO Doc 7300.

\bibitem{beer}
Beer, ``Bestimmung der absorption des rothen lichts in farbigen
  fl{\"u}ssigkeiten,'' \emph{Annalen der Physik}, vol. 162, no.~5, pp. 78--88,
  1852.

\bibitem{awos}
F.~A. Circular, ``Automated weather observing systems (awos) for non-federal
  applications,'' 3 2017.

\bibitem{matthews2023veia}
M.~P. Matthews, ``Visibility estimation through image analytics,'' MIT Lincoln
  Laboratory, Tech. Rep. ATC-453, 2023.

\bibitem{li}
S.~Li, H.~Fu, and W.-L. Lo, ``Meteorological visibility evaluation on webcam
  weather image using deep learning features,'' \emph{Int. J. Comput. Theory
  Eng}, vol.~9, no.~6, pp. 455--461, 2017.

\bibitem{RMEP}
P.~Su, Y.~Liu, S.~Tarkoma, A.~Rebeiro-Hargrave, T.~Petäjä, M.~Kulmala, and
  P.~Pellikka, ``Retrieval of multiple atmospheric environmental parameters
  from images with deep learning,'' \emph{IEEE Geoscience and Remote Sensing
  Letters}, vol.~19, pp. 1--5, 2022.

\bibitem{visnet}
A.~Palvanov and Y.~I. Cho, ``Visnet: Deep convolutional neural networks for
  forecasting atmospheric visibility,'' \emph{Sensors}, vol.~19, no.~6, p.
  1343, 2019.

\bibitem{kipfer2017fog}
K.~Kipfer, ``Fog prediction with deep neural networks,'' Master's thesis, ETH
  Zurich, 2017.

\bibitem{simonyan2014very}
K.~Simonyan and A.~Zisserman, ``Very deep convolutional networks for
  large-scale image recognition,'' \emph{arXiv preprint arXiv:1409.1556}, 2014.

\bibitem{koonce2021resnet}
B.~Koonce, ``Resnet 34,'' \emph{Convolutional neural networks with swift for
  tensorflow: image recognition and dataset categorization}, pp. 51--61, 2021.

\bibitem{SDV18}
\BIBentryALTinterwordspacing
C.~Sakaridis, D.~Dai, and L.~Van~Gool, ``Semantic foggy scene understanding
  with synthetic data,'' \emph{International Journal of Computer Vision}, vol.
  126, no.~9, pp. 973--992, Sep 2018. [Online]. Available:
  \url{https://doi.org/10.1007/s11263-018-1072-8}
\BIBentrySTDinterwordspacing

\bibitem{xie2024synfog}
Y.~Xie, H.~Wei, Z.~Liu, X.~Wang, and X.~Ji, ``Synfog: A photo-realistic
  synthetic fog dataset based on end-to-end imaging simulation for advancing
  real-world defogging in autonomous driving,'' in \emph{Proceedings of the
  IEEE/CVF Conference on Computer Vision and Pattern Recognition}, 2024, pp.
  21\,763--21\,772.

\bibitem{belaroussi2014impact}
R.~Belaroussi and D.~Gruyer, ``Impact of reduced visibility from fog on traffic
  sign detection,'' in \emph{2014 IEEE intelligent vehicles symposium
  proceedings}.\hskip 1em plus 0.5em minus 0.4em\relax IEEE, 2014, pp.
  1302--1306.

\bibitem{schaupp2025synthetic}
J.~Schaupp, C.~Mourning, and J.~Murray, ``Synthetic imagery in the use of
  training visibility estimators,'' in \emph{AIAA SCITECH 2025 Forum}, 2025, p.
  2325.

\bibitem{yang2023multi}
W.~Yang, Y.~Zhao, Q.~Li, F.~Zhu, and Y.~Su, ``Multi visual feature fusion based
  fog visibility estimation for expressway surveillance using deep learning
  network,'' \emph{Expert Systems with Applications}, vol. 234, p. 121151,
  2023.

\bibitem{gated2depth2019}
T.~Gruber, F.~Julca-Aguilar, M.~Bijelic, and F.~Heide, ``Gated2depth: Real-time
  dense lidar from gated images,'' in \emph{The IEEE International Conference
  on Computer Vision (ICCV)}, 2019.

\bibitem{jacobs07amos}
N.~Jacobs, N.~Roman, and R.~Pless, ``Consistent temporal variations in many
  outdoor scenes,'' in \emph{IEEE Conference on Computer Vision and Pattern
  Recognition (CVPR)}, Jun. 2007, pp. 1--6, acceptance rate: 23.4\%.

\bibitem{mourning2024towards}
C.~Mourning, D.~Wright, and J.~Murray, ``Towards a low cost distributed awos:
  Machine learning for cost effective visibility estimation,'' in \emph{AIAA
  SCITECH 2024 Forum}, 2024, p. 2355.

\bibitem{ASTM_F3673-23}
{ASTM International}, ``Standard specification for performance for weather
  information reports, data interfaces, and weather information providers
  (wips),'' ASTM Standard F3673-23, ASTM International, West Conshohocken, PA,
  2024, available from: \url{https://www.astm.org/Standards/F3673.htm}.

\end{thebibliography}

%

\begin{IEEEbiography}[{\includegraphics[height=1.25in,clip,keepaspectratio]{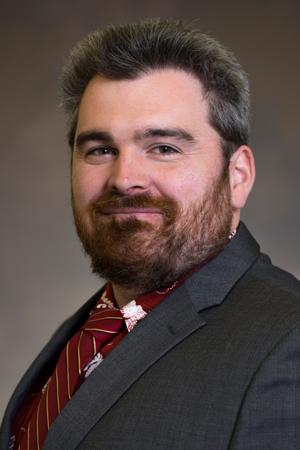}}]{Chad Mourning}
Chad Mourning is Principal Investigator of the VizSim Lab at Ohio University.  Chad Mourning is an Assistant Professor of Computer Science at Ohio University.  Chad is a native of Middleport, Ohio and received his Ph.D. from Ohio University in 2015.  Chad's main research areas include Computer Graphics \& Data Visualization, Modeling \& Simulation, Augmented \& Virtual Reality, and Machine Learning with applications in Meteorology.  Chad is currently serving as Principal Investigator on the Low Altitude Weather Network project and has other ongoing funded projects by NASA, the Ohio Space Grant Consortium, and the FAA.
\end{IEEEbiography}

\begin{IEEEbiography}[{\includegraphics[height=1.25in,clip,keepaspectratio]{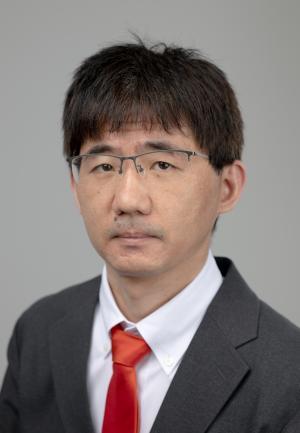}}]{Zhewei Wang}
Zhewei Wang is a Visiting Assistant Professor of Artificial Intelligence in the School of Electrical Engineering and Computer Science at Ohio University. He completed his Ph.D. in Computer Science and M.S. in Biomedical Engineering at Ohio University in 2020 and 2019, respectively. From 2021 to 2023, he conducted research at Massachusetts General Hospital of Harvard Medical School. His research interests include Deep Learning, Machine Learning, Computer Vision, Medical Image Analysis, Natural Language Processing, Graph Neural Networks, and Reinforcement Learning.\end{IEEEbiography}


\begin{IEEEbiographynophoto}{Justin Murray}
    

Justin received a Bachelor of Science in Computer Science from Ohio University in 2023 and is currently pursuing a Master of Science in Computer Science from Ohio University.  Justin is a 2023 recipient of the Ohio Space Grant Consortium Research Fellowship.  Current research interests include machine learning, and computer graphics. 
\end{IEEEbiographynophoto}




\end{document}